# Does Syntax Matter? A Graph-Augmented Variational Topic Model for Computational Social Sciences

*Alessandro Meneghini – University of Udine – alessandro.meneghini@uniud.it*

## Abstract

Topic modeling is a widely used methodology in computational social sciences. Traditional approaches rely on Bag-of-Words representations and generative probabilistic models (LDA, STM), while recent clustering-based architectures such as BERTopic operate through clustering of word embeddings. This paper introduces the Structural Contextual Probabilistic Topic Model (SCPTM), an architecture that encodes syntactic dependency relations directly into the topic inference process. SCPTM models a corpus as a heterogeneous graph where documents and words are connected by lexical co-occurrence edges and dependency-parsed syntactic edges. A Graph Attention Network processes this structure within a Variational Autoencoder framework to produce probabilistic, mixed-membership topic distributions.

We evaluate seven topic modeling techniques, including four SCPTM ablations isolating the contribution of lexical and syntactic graph edges, across four corpora differing in register and discourse structure. Using an evaluation framework that includes standard coherence metrics (C_V, C_NPMI), topic quality measures (diversity, intra-topic concentration, NMI), and diagnostics quantifying sentiment differences in topic descriptors, we investigate whether and under what corpus conditions syntactic augmentation provides measurable benefits.

Results indicate that SCPTM's neural architecture yields substantial gains in document-topic alignment over generative baselines, but that these gains are attributable to the variational encoder rather than to syntactic edges. The specific contribution of syntax emerges instead in topic diversity, where graph-augmented variants outperform the no-graph baseline across all corpora, and in the qualitative character of extracted descriptors: dependency paths capture predicate-argument structures and evaluative stance in deliberative registers, while proving redundant in technical and institutional corpora. The valence gap between multi-word and unigram descriptors is positive for all SCPTM variants, but its magnitude is driven primarily by phrase grouping rather than syntactic filtering. We conclude that syntactic encoding matters conditionally: it provides measurable benefits when textual meaning relies on action, agency, and argumentative stance, and introduces structural noise when register is primarily informational or administrative.

## 1. Introduction

The computational analysis of large-scale textual corpora has become a more and more widely used procedure across the social sciences, from political discourse analysis and policy evaluation to the study of online communities and public opinion (Chen et al., 2023; Grimmer & Stewart, 2013; Lazer et al., 2020). Specifically, topic modeling is a technique that lies in the broader field of quantitative text analysis (Tuzzi, 2018), and is used for unsupervised identification of latent themes in a collection of documents, enabling researchers to map the themes of wide corpora of texts without manual annotation. The scale and ambition of contemporary research often pushes researchers toward methods that trade interpretive depth for breadth of coverage. Large language models, in particular, have been shown to homogenise linguistic and semantic features, converging toward specific patterns according to their training data at the expense of more nuanced positions (Sourati et al., 2025), similarly to risks long documented for large survey research (Jackson, 2026). Quantitative text analysis techniques, and specifically topic modeling, should resist this trade-off, and even when applied at scale, they should preserve and describe the different facets of meaning, returning an overview of a corpus' themes while sacrificing as little as possible of their underlying complexity, in a "distant reading" manner (Moretti, 2013).

Established approaches to topic modeling describe themes through ranked lists of words which are relevant for each topic. Latent Dirichlet Allocation (Blei, 2003), the technique that pioneered topic modeling, remains a widely used approach for exploring topics, due to its statistical interpretability. In LDA documents are considered as a mixture of topics, which are at the same time considered as a mixture of words. On the other hand, word embedding-based approaches such as Top2Vec (Angelov, 2020), CombinedTM (Bianchi et al., 2021) and BERTopic (Grootendorst, 2022) substantially improve the semantic quality of retrieved topics, by capturing the semantic meaning thanks to pre-trained transformers models.

This paper introduces the Structural Contextual Probabilistic Topic Model (SCPTM), a hybrid architecture that encodes semantic features and syntactic dependency relations directly into the topic inference process. Its central aim is to determine when syntactic encoding improves topic quality, and under what corpus conditions. This question is motivated by the observation that different textual registers encode meaning differently. In deliberative or argumentative texts, agency, stance, and evaluative positioning are grammaticalized through predicate-argument structures (e.g., "protect border" vs. "border communities"). In technical or administrative texts, by contrast, syntax offers little beyond what lexical co-occurrence already captures. The methodological contribution of this paper is twofold. First, the introduction of SCPTM as an architecture that combines syntactic graph encoding with probabilistic mixed-membership topic inference. Second, we propose a comparative evaluation framework designed to answer the question on the impact of syntax for topic modeling by assessing whether, where, and under what conditions syntactic encoding yields measurable benefits. The framework operates along three interconnected dimensions: model comparison, metric heterogeneity, and syntactic diagnostics. On the first dimension, we evaluate seven models (LDA, CombinedTM, four SCPTM ablations varying graph compositions, and BERTopic) across four corpora that differ in register, formality, and discourse structure: the 20 Newsgroups benchmark, the European Parliament Debates corpus, a Reddit political dataset, and a hate speech corpus. On the second dimension, we deploy a battery of metrics that capture different facets of topic quality: standard coherence (C_V, C_NPMI), topic diversity, intra-topic concentration, and label alignment (NMI). Because these metrics measure fundamentally different properties (semantic coherence, lexical spread, quality of document clustering, and correspondence to external categories) no single metric is sufficient to answer. On the third dimension, for SCPTM variants we introduce Multi-Word Expression specific diagnostics. These include complementarity, and the valence gap, which quantify the distinctive properties of syntactically based collocations relative to unigram baselines. This evaluation framework explores under which conditions syntactic encoding provides added value.

The remainder of this paper is organised as follows. Section 2 reviews existing definitions of topics and positions SCPTM within the literature. Section 3 reviews related work. Section 4 describes the model architecture and training procedure. Section 5 presents the experimental setup, evaluation framework, and results, including cross-corpus comparisons. Section 6 discusses implications and limitations, and Section 7 concludes.

## 2. What Is a Topic? A Matter of Representation

In the context of this paper, before presenting the model architecture, it is useful to review how different computational paradigms conceptualize and operationalize topics, as these are design choices that determine what a model can and cannot grasp from text.

### 2.1 Topic as Probability Distribution: Generative Models

In the generative probabilistic tradition started by Latent Dirichlet Allocation (Blei, 2003), a topic is defined as a probability distribution over a fixed vocabulary:

$$\beta_k \in \Delta^V$$

where high-probability words characterize the topic. A document is a mixture over K such distributions:

$$\theta_n \in \Delta^K$$

The generative process assumes that each word is drawn by first sampling a topic from $\theta_n$ and then sampling a word from the corresponding $\beta_k$. This conceptualization of topic has two implications. First, topics are defined relationally: they emerge from the co-occurrence structure of the corpus rather than from any external semantic annotation. Second, documents have mixed membership: a single text can be associated with multiple topics, with each topic receiving a probability weight, and this can be helpful where the difference between concepts is nuanced and contextual for example for example in social and political texts that address multiple themes simultaneously (DiMaggio et al., 2013; Roberts et al., 2014). Topic modeling techniques relying in the LDA approach are based on Bag-of-Words (BoW) representations of text, which treat each document as a set of its tokens, "disregarding grammatical and syntactical roles but keeping multiplicity" (Misuraca & Spano, 2020, p. 18). Consequently, two sentences such as "you made me feel loved" and "I made you feel loved", which share the same vocabulary but differ in who is the subject of the action, produce identical or highly similar representations. More generally, BoW models cannot distinguish between different senses of the same word ("bank" as a financial institution vs. "bank" as a river edge), nor can they recognize that different words may refer to the same concept ("river" and "stream" are treated as independent vocabulary items).

### 2.2 The Word Embeddings Perspective

The introduction of static and dynamic word embedding representations, such as CBOW (Mikolov et al., 2013), BERT (Devlin et al., 2019) and SBERT (Reimers & Gurevych, 2019), allowed a change in perspective on "what" to define as a topic. In embedding-based models such as Top2Vec (Angelov, 2020) and BERTopic (Grootendorst, 2022), a topic is a region in a continuous semantic space, identified by a clustering algorithm, for example in BERTopic density-based clustering techniques such as HDBSCAN (McInnes et al., 2017).

This approach addresses two limitations of BoW-based models. First, it resolves the synonymy problem: semantically related terms such as "doctor" and "physician" are geometrically close in the embedding space and will be grouped together even without frequent co-occurrence. Second, transformer-based embeddings provide some degree of sense disambiguation, as word representations are context-dependent (Devlin et al., 2019). The resulting topics tend to exhibit higher semantic coherence than BoW-based models (Egger & Yu, 2022).

However, this approach poses different limitations. Clustering algorithms, whether density-based (HDBSCAN) or distance-based (K-Means), produce hard assignments of documents to clusters. This is appropriate for applications where documents are expected to address a main dominant theme, but it might not work adequately in mixed-membership scenarios. A further consideration is that, while document representations are enriched by transformer-based encoding, the extraction of topic keywords in these models typically reverts to frequency-based metrics such as c-TF-IDF, implicitly re-adopting a BoW logic at the output stage (Egger & Yu, 2022). This means that the semantic richness of the input representation might not be fully preserved in the final topic description. Lastly, the representation generated by pretrained language models introduces a limitation in interpretability. While these models produce semantically rich embeddings, they function as black boxes as the dimensions of the embedding space lack direct interpretation. As a result, researchers cannot easily explain why a set of documents is assigned to the same or to different clusters, thus exposing a significant obstacle to the responsible deployment of pretrained models in critical domains (Bommasani et al., 2022).

### 2.3 Topics as Structural Encoding of Semantics: The SCPTM Architecture

The two approaches previously described offer complementary strengths, being the explainability of the model, or the semantic richness captured through BERT. The approach taken in this paper combines elements of both paradigms while adding a third dimension: syntactic structure. Rather than defining a topic solely by the words that appear in it (as in LDA) or by its position in a semantic space (as in BERTopic), SCPTM defines a topic as a recurring pattern of syntactic and semantic dependencies, configurations in which specific types of heads, dependents, and dependency relations (e.g., nominal subjects, direct objects, adjectival modifiers, noun modifiers, noun compounds, conjunctions, clausal complements) appear together across a set of documents.

What does this mean in practice? Consider a corpus of political discourse. A word like "border" may appear in different syntactic contexts – as the object of "protect" or "control" in nationalist discourse, or as a modifier of "communities" or "families" in humanitarian discourse. In a Bag-of-Words model, this semantic similarity is not treated, the word "border" contributes the same signal regardless of its role. In SCPTM, the dependency structure is preserved, as the model propagates information along syntactic edges, so the representation of a word such as "border" is updated differently depending on whether it appears as an object ("protect border"), a modifier ("border communities"), or a compound ("border control"). As a result, two documents that share vocabulary may receive different topic assignments if their vocabulary is used in systematically different syntactic configurations. This conceptual shift has practical consequences for topic interpretation. In generative models, topics are described by ranked lists of unigram keywords. In SCPTM, topics are described by both unigrams and multi-word expressions (MWEs). SCPTM extracts MWEs from the dependency graph as triples of the form (head, relation, dependent). Again, "protect border" becomes (protect, obj, border), where obj indicates that "border" is the direct object of "protect". This preserves the syntactic relationship between the two words.

SCPTM starts from representing texts as a heterogeneous graph: documents and words are nodes, connected by lexical co-occurrence edges and syntactic dependency edges. The model encodes this graph through a Graph Attention Network (GAT; Veličković et al., 2018), which propagates information along both types of edges, producing a vector representation for each document that reflects both its lexical content and the syntactic roles of its words. These document representations are then fed into a Variational Autoencoder (VAE; (Kingma & Welling, 2022), which infers a probability distribution over topics for each document. The VAE is chosen because it produces mixed-membership topic assignments, while its probabilistic formulation provides a latent space that prevents overfitting to individual documents. This combination preserves the interpretability of generative models like LDA while incorporating the structural information of graph-based approaches. Finally, topic representations are extracted as both unigrams and multi-word expressions (MWEs) grounded in the dependency graph. The technical details of these components are described in Section 4.

## 3. Related Work

SCPTM defines a topic as a recurring pattern of syntactic dependencies, configurations in which specific types of dependency relations appear together across a set of documents (see Section 2.3). This definition has three architectural implications: the model must operate on syntactic structure, preserve mixed-membership, and capture semantic similarity between words.

### 3.1 Syntactic extensions of topic models

The work that most directly anticipates the integration of syntax and topic modeling is that of Wallach (2006), which extends LDA to distributions over bigrams to capture local dependencies between words. Boyd-Graber & Blei (2009) introduce Syntactic Topic Models, in which dependency structure is incorporated into LDA's generative process: each word in a syntactic tree is generated from a topic that depends both on the document's topic mixture and on the local syntactic context (whether the word is a head, a dependent, and what relation links them). All these approaches share the intuition that grammatical information contains useful signal for topic modeling, implement it within the generative bag-of-words paradigm.

### 3.2 Bag-of-words approaches

The probabilistic tradition initiated by LDA (Blei, 2003) considers each document as a distribution over topics and each topic as a distribution over the vocabulary. Extensions such as the Correlated Topic Model (CTM; Blei & Lafferty, 2007) and the Structural Topic Model (STM; Roberts et al., 2014) retain the same mixed-membership logic, introducing respectively topic correlations and covariates. The strength of these models for the social sciences lies in their ability to represent texts that simultaneously address multiple topics with different weights. Their limitation, common to all bag-of-words models, is the inability to distinguish different uses of the same word across different syntactic contexts.

### 3.3 Embedding-based approaches

The use of distributed text representations has led to a redefinition of the concept of topic. In the Embedded Topic Model (ETM; Dieng et al., 2020) words and topics are projected into a shared vector space, where geometric proximity encodes semantic similarity. The Combined Topic Model (CombinedTM; Bianchi et al., 2021) uses contextual SBERT embeddings in a VAE architecture, replacing the bag-of-words input with dense document representations. BERTopic (Grootendorst, 2022) and Top2Vec (Angelov, 2020) shift from the topic as a distribution of words entirely: documents are represented into an embedding space, then they are clustered after dimensionality reduction, and topics are described as frequency-based metrics such as c-TF-IDF.

### 3.4 Graph-based approaches

A narrower approach employs graphs for topic modeling. TextGCN (Yao et al., 2019) builds a heterogeneous document-word graph with edges weighted by Pointwise Mutual Information (PMI) and uses Graph Convolutional Networks (GCNs) for text classification: the graph propagates information between documents and words by using global co-occurrence. PMI connects words that co-occur across the entire corpus, not just within the same document, creating a global word–word network. This means two documents that share no vocabulary can still exchange information during training if they are connected through intermediate words that appear with both, for instance, even if "migration" and "asylum" never appear together in one document, if both co-occur with "border" somewhere in the corpus, the graph links them via that shared neighbor. TextGCN uses this network system for semi-supervised classification with few labels. Sayyadi and Raschid apply community detection algorithms to lexical co-occurrence networks to extract more stable topics (Sayyadi & Raschid, 2013). The graph structure makes topic extraction less sensitive to the frequency skew and initialization randomness that can cause LDA topics to collapse onto generic high-frequency terms or to shift across runs; a word's membership depends on its network neighborhood rather than on per-topic multinomials.

## 4. Structure and Semantics: the SCPTM Model

SCPTM operates in three stages (Figure 1). First, it constructs a representation of the corpus that captures both semantic content and syntactic structure, by embedding words and documents with SBERT and by dependency-parsing each document. Second, it encodes this dual representation into a topic distribution for each document, using a graph attention network followed by a variational autoencoder. Third, it extracts topic descriptors in two forms: single words and multi-word expressions grounded in syntactic dependency paths.

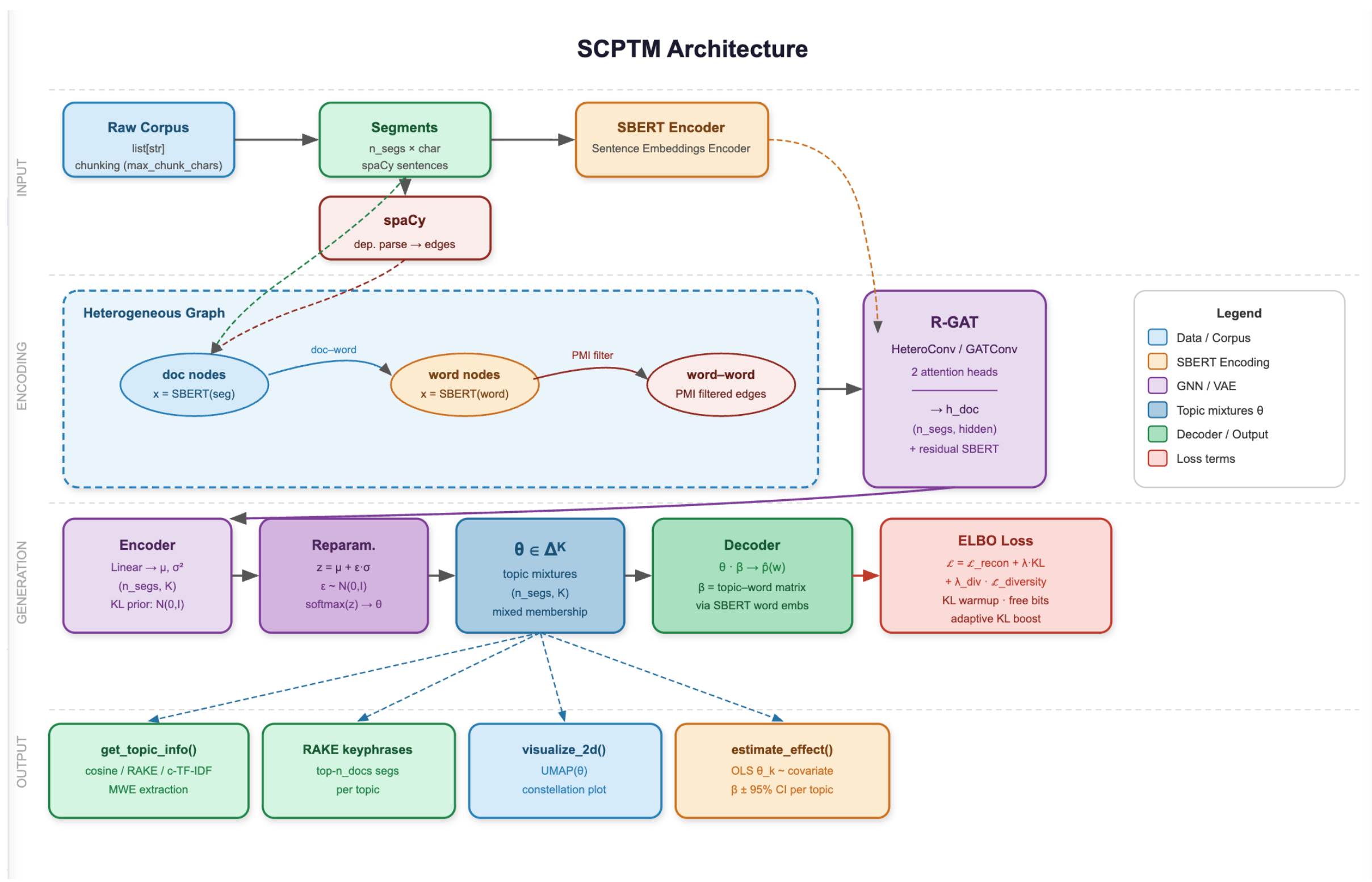

*Figure 1- The SCPTM Architecture, divided by step: input, encoding, model generation, output*

### 4.1 Semantic Initialization

Before any graph structure is introduced, SCPTM needs a semantic representation of words, documents, and topics as a starting point.

Let $D$ = {*d*$_1$, …, *d*_N} be a corpus of $N$ documents and $V$ = {*w*$_1$, …, *w*_L} be the vocabulary of unique lemmas extracted after removing stop words. Each document is embedded using SBERT, producing a single vector that captures its overall meaning in context. For individual words, the situation is more delicate. The same lemma (e.g., "bank") can appear in multiple contexts with different meanings ("river bank" vs. "financial bank"). To obtain a single embedding for each lemma, we average the SBERT vectors of all its occurrences in the corpus. The resulting vector represents the typical contextual meaning of that lemma: words that tend to appear in similar contexts end up close to one another in the embedding space, while words used in sharply different ways are pulled apart. Topic embeddings {$\alpha_1$, …, α_K} are initialized from the $K$ centroids of a k-means clustering of the word embedding space. In SBERT space, semantically related words (e.g., *carbon, pollution, emissions*) form dense clusters. Placing a topic centroid at the centre of one such cluster gives the model an informed starting point: topic *k* appears close to a semantically coherent set of words. These centroids are learnable parameters and move during training according to two criteria: the reconstruction loss pulls them toward words that co-occur in documents, while the topic diversity penalty pushes apart centroids that become too similar. The clustering step therefore supplies an initial configuration that is both semantically structured and malleable enough to be reshaped by the data.

### 4.2 Syntactic Graph Construction

The corpus is modelled as a global heterogeneous graph $G = (V^G, E^G)$ where $V^G = D \cup V$. The edge set consists of two different relation types.

- Document-word edges (contains): A directed edge connects d_i to w_j if lemma w_j appears in document d_i above a minimum document-frequency threshold. These edges preserve lexical composition information, which words appear in which documents.
- Word-word edges (relates): Directed edges between word nodes are drawn exclusively from syntactic dependency parses produced by spaCy's pipeline (Honnibal et al., 2020). We retain a fixed subset of Universal Dependencies relation labels (Nivre et al., 2020) that typically connect content words (nsubj, obj/dobj, amod, nmod, compound, conj, xcomp) following the standard content/function-word distinction in dependency grammar. Relations that mark grammatical role without contributing independent lexical content (det, punct, aux, among others) are excluded.

The key consequence of this design is that the same word can receive different representations depending on its syntactic context. Consider again the word "border" in a corpus of political discourse. It may appear as the direct object of "protect" in one document ("protect border") and as a noun modifier in another ("border communities"). In the syntactic graph, "border" receives different messages along a *dobj* edge from "protect" than along an *nmod* edge to "communities". After aggregation (described in the next section), the representation of "border" encodes the typical syntactic roles it occupies, and the document representation inherits this differentiation. Two documents that share vocabulary may therefore receive different topic assignments if their vocabulary is used in systematically different syntactic configurations.

### 4.3 Graph Attention Encoder

The graph constructed in Section 4.2 contains two types of nodes: documents and words connected by two types of edges: document–word edges ("contains") and word–word edges ("relates"). The next step is to transform this graph into a vector representation for each document. SCPTM uses a Graph Attention Network (GAT; **(Veličković et al., 2018)** for this task. Each node updates its representation by "attending" to its neighbours, weighting some neighbours more than others depending on their relevance. For example, a word like "border" will be updated differently depending on whether it appears as a direct object in "protect border" or as a noun modifier in "border communities." The attention mechanism learns to weigh these syntactic contexts according to their relevance to the document's topic.

For each edge type, the model computes an attention score between a node *i* and each of its neighbours *j*:

$$\alpha_{ij} = \text{softmax}_{(j \in N_i)} (a^T [Wh_i \| Wh_j])$$

Here $h_i$ is the current representation of node *i*, $W$ and *a* are learned parameters, ‖ denotes concatenation, and $N_i$ is the set of nodes connected to *i* by an edge of the relevant type. The updated representation of node *i* is a weighted sum of its neighbours' transformed representations:

$$h_i = \sum_{(j \in N_i)} \alpha_{ij} Wh_j$$

Each node's new representation is a weighted average of its neighbours' representations, where the weights are learned and context-dependent. A deliberate design choice is that all retained word–word dependency edges are pooled into a single relation type ("relates") and processed by one shared attention mechanism. We do not learn separate attention parameters for each dependency label (nsubj, dobj, amod, etc.). The model's selectivity with respect to syntax is exercised at the graph-construction stage (Section 4.2), where only informative dependency labels are retained as edges, rather than through relation-specific weights inside the attention layer. This design keeps the parameter count manageable and avoids overfitting to sparse dependency labels, but it means the model cannot learn that dobj relations are systematically more informative than compound relations. Extending the architecture to learn relation-specific attention weights is a natural direction for future work.

The same attention mechanism is applied to the document–word ("contains") edges and their reverse ("rev_contains"): each word node also receives messages from the documents that contain it, and each document node receives messages from the words it contains. Messages arriving via different incoming edge types are combined by mean pooling. Because all edge types are computed from the same input node features within this single layer, word- and document-side updates happen in parallel.

The word→document message is combined with the document's own SBERT embedding through a residual connection:

$$h_d = Linear(SBERT(d)) + g \odot Message(\text{word} \rightarrow \text{doc})(\text{d})$$

where $g$ is a learnable, sigmoid-bounded gate initialized close to zero. This design was motivated by a failure mode observed during development, in which an unweighted merge of the two signals allowed word-level aggregation to over-smooth document representations and degrade downstream clustering quality. The gate lets training determine, per corpus, how much of the graph-derived signal is useful, while guaranteeing that the document's own semantic content is never discarded outright. At the end of this stage, each document has a vector representation $h_d$ that combines its own semantic content (from SBERT) with information from its constituent words, which in turn have been enriched by their syntactic neighbours. This representation is the input to the next stage.

**4.4 From Document Representation to Topic Distribution**

The document representation $h_d$ produced in Section 4.3 is a vector that summarises the document's content. To obtain a topic distribution (a set of $K$ numbers that sum to one and indicate the document's engagement with each topic) SCPTM uses a Variational Autoencoder (VAE; Kingma & Welling, 2022).The VAE is chosen because it produces mixed-membership topic assignments, so each document receives a continuous probability distribution over topics, rather than a single hard cluster. The probabilistic formulation also provides a regularised latent space that prevents overfitting to individual documents.
The document representation $h_d$ is passed through two linear transformations to produce the parameters of a Gaussian distribution:

$$\mu_d = W_\mu \, h_d + b_\mu, \;\; \log \sigma_d^2 = W_\sigma \, h_d + b_\sigma$$

The distribution $q(z|d) = N(\mu_d, \sigma_d\text{^}2\, I)$ represents the model's uncertainty about the document's topic proportions. To obtain a concrete topic mixture, the model draws a sample using the reparameterization trick:

$$z_d = \mu_d + \sigma_d \odot \varepsilon, \varepsilon \sim N(0, I)$$

and applies a softmax:

$$\theta_d = \text{softmax}(z_d) \in \Delta\text{^}K$$

The vector $\theta_d$ is the document's topic distribution, a set of positive numbers that sum to one, directly interpretable as the proportions with which the document addresses each of the $K$ topics.

The model is trained by maximizing the Evidence Lower Bound (ELBO). For a document $d$, the ELBO is:

$$L_d = E_{q(z|d)} [\log(p(d|z))] - \beta \cdot D_{KL} (q(z|d) \parallel p(z))$$

The first term measures how well the sampled topic mixture $\theta_d$ can reconstruct the document's actual word counts. The model assumes that words are generated from the topic mixture via the topic–word distribution $\beta$, computed from the topic embeddings $\alpha_k$ and word embeddings $e_w$:

$$\beta_{kv} = \text{softmax}_v(\cos(\alpha_k, e_w)/\tau)$$

Here $\cos(\alpha_k, e_w)$ is the cosine similarity between the topic embedding and the word embedding, and $\tau = 0.1$ is a temperature parameter that sharpens the distribution. In plain terms: if the topic mixture is accurate, the model should assign high probability to the words that actually appear in the document. The second term penalizes the learned posterior $q(z|d)$ for straying too far from a standard Gaussian prior $p(z) = N(0, I)$. This prevents the model from memorising each document's exact word counts and forces it to find regularities that generalise across documents. It also keeps $\sigma_d$ bounded away from zero, preventing the posterior from collapsing to a point mass.

Additionally, a topic diversity penalty encourages the topic embeddings to remain distinct from one another:

$$\ell_{div} = 1/(K(K-1)) \sum_{(j \neq k)} \cos(\alpha_j, \alpha_k)$$

This discourages the model from collapsing multiple topics into the same semantic region.

### 4.5 Inference and Topic Extraction

At inference, document topic mixtures are obtained by passing the graph through the trained encoder and taking $\theta_d = \text{softmax}(\mu_d)$ without sampling. Topic keywords are extracted by ranking vocabulary terms by cosine similarity to each topic embedding $\alpha_k$ in the shared embedding space: the top-n terms for topic *k* are the vocabulary items whose SBERT embeddings are most geometrically proximate to the topic centroid.

Beyond single-word keywords, SCPTM extracts multi-word expression (MWE) descriptors for each topic. MWEs are identified as directed paths of length 2 in the syntactic graph: starting from any content word node, the algorithm traverses one of the eight retained dependency relation types (nsubj, obj, amod, nmod, compound, conj, xcomp) to reach a dependent word, forming a (head, relation, dependent) triple. Each candidate MWE is scored by the mean cosine similarity of its constituent SBERT word embeddings to the topic embedding $\alpha_k$. The top-20 MWEs per topic are retained as phrase-level descriptors.

## 5. Empirical Evaluation

As stated in Section 1, this paper's empirical contribution consists of a comparative evaluation framework operating along three interconnected dimensions, which the following subsections take up in turn. Model comparison (Section 5.3, Table 1) situates SCPTM's four ablations against LDA, CombinedTM, and BERTopic on standard topic-quality metrics, across four corpora chosen to vary in register, formality, and discourse structure. For what concerns SCPTM, we evaluate four configurations of the architecture, designed to isolate the contribution of lexical and syntactic graph edges. *SCPTM-none* removes the graph entirely: the variational autoencoder receives only the document's SBERT embedding, making it similar to CombinedTM with KL annealing and k-means topic initialization. *SCPTM-nosyn* retains only the document–word bipartite edges, allowing messages to propagate between documents and their constituent words but not between words themselves. *SCPTM* in its default configuration adds word–word edges drawn from the eight informative dependency relation types described in Section 4.2 (*nsubj, obj, amod, nmod, compound, conj, xcomp*). *SCPTM-fulldep* is identical to SCPTM except that word–word edges are drawn from all dependency relation types that connect content words, without restricting to the eight informative types. Function-word relations (*det, punct, aux*) remain excluded in all variants. The difference between SCPTM and SCPTM-fulldep therefore isolates the effect of the syntactic filter itself. Metric heterogeneity (Section 5.2) reflects the position that no single score can adjudicate topic quality on its own, particularly for corpora without an external category system to validate against. We therefore report multiple complementary diagnostics: coherence (C_V, C_NPMI), topic diversity, clustering-label alignment (NMI), and for the SCPTM variants phrase-level metrics including complementarity and the valence gap. Syntactic diagnostics (Section 5.4, Table 2) then isolates what, if anything, the syntactic graph specifically contributes: specificity, complementarity, and valence gap compare SCPTM's dependency-grounded multi-word descriptors against unigram baselines and against a generic bigram baseline computed for the non-syntactic models. Together these three dimensions are designed to locate the conditions under which syntactic encoding helps.

All experiments were run on a single virtualized server equipped with an NVIDIA A40 GPU (48GB VRAM), 8 vCPUs (Intel Xeon Gold 6342 @ 2.80GHz), and 31GB RAM. The full sweep across four corpora, seven models, the topic-count sweep, and three random seeds required approximately 11,5 hours of wall-clock compute time.

### 5.1 Corpora Selection

We evaluate on four English-language corpora chosen to span a range of discursive registers and degrees of political polarization, allowing us to test whether the benefit of syntactic structure is corpus-conditional rather than universal.

- 20 Newsgroups (Mitchell, 1997) is a standard topic-modelling benchmark of roughly 18,000 Usenet posts spanning twenty largely non-political subject areas (e.g., computing, sports, religion, tech). Its informal register and topical organization make it a useful baseline against which the remaining, more polarized corpora can be contrasted; we fix K=20 for this corpus to match its known ground-truth category structure.
- EU Parliament Debates (Chalkidis & Brandl, 2024) consists of formal plenary speeches delivered by Members of the European Parliament, filtered here to English-native speakers. Its register is deliberative and institutional.
- Reddit Politics (Gajare, 2024) comprises roughly 13,000 posts drawn from Liberal- and Conservative-leaning subreddits, combining post titles with body text where available. Its register is informal.
- Measuring Hate Speech (Sachdeva et al., 2022) is a corpus of social-media text annotated for hate-speech severity using Rasch measurement theory; we use the raw text without the severity labels.

For the three non-20NG corpora we sweep K {5, 10, 15, 20, 25, 30}; the Hate Speech corpus is additionally evaluated at K=8.

### 5.2 Evaluation Metrics

Diagnostics for topic-model quality can be quite heterogeneous. We follow the position articulated by Sciandra, Trevisani, and Tuzzi (2023), who argue that meaningful topic-model evaluation requires triangulating across multiple, partially conflicting quantitative signals rather than optimizing a single score, a stance we adopt explicitly given how differently the four corpora above are structured. Our evaluation therefore combines standard coherence measures (C_V, C_NPMI) for comparability with prior work, topic diversity to assess lexical spread, normalized mutual information (NMI) to measure alignment with external ground-truth labels where available, as well as phrase-level diagnostics for SCPTM variants: complementarity and the valence gap, which assess the distinctive properties of syntactically grounded multi-word descriptors relative to unigram baselines.

### 5.3 Baseline Comparison: Neural Architecture vs. Generative Models

This section presents the results of the evaluation metrics related to the topic quality, which are reported in Table 1.

*Table 1. Topic quality metrics by model and corpus, evaluated at each model's best K (mean over 3 seeds).*

| Corpus | Model | Best K | C_V | C_NPMI | Diversity | NMI |
|---|---|---|---|---|---|---|
| **20NG** | BERTopic | 20 | **0.5842** | **0.0709** | **0.9295** | **0.4859** |
| | CTM | 20 | 0.5234 | -0.2360 | 0.9219 | 0.2315 |
| | LDA | 20 | 0.5318 | 0.0327 | 0.4383 | 0.1484 |
| | SCPTM | 20 | 0.3930 | -0.2792 | 0.7450 | 0.3512 |
| | SCPTM-fulldep | 20 | 0.3881 | -0.2649 | 0.7483 | 0.3403 |
| | SCPTM-none | 20 | 0.4028 | -0.2520 | 0.7383 | 0.3911 |
| | SCPTM-nosyn | 20 | 0.3869 | -0.2736 | 0.7750 | 0.3512 |
| **EU_Debates** | BERTopic | 10 | **0.6094** | **0.1496** | 0.8741 | — |
| | CTM | 30 | 0.5408 | -0.0083 | **0.9170** | — |
| | LDA | 25 | 0.4683 | 0.0332 | 0.3427 | — |
| | SCPTM | 30 | 0.3709 | -0.2142 | 0.5256 | — |
| | SCPTM-fulldep | 30 | 0.3718 | -0.2098 | 0.5244 | — |
| | SCPTM-none | 30 | 0.3740 | -0.2012 | 0.4067 | — |
| | SCPTM-nosyn | 30 | 0.3626 | -0.2185 | 0.5344 | — |
| **HateSpeech** | BERTopic | 15 | 0.5622 | **0.0711** | 0.7619 | **0.3113** |
| | CTM | 15 | **0.5773** | 0.0126 | **0.9795** | 0.1138 |
| | LDA | 15 | 0.3609 | -0.0041 | 0.3089 | 0.0160 |
| | SCPTM | 30 | 0.4009 | -0.3045 | 0.6922 | 0.2131 |
| | SCPTM-fulldep | 30 | 0.4124 | -0.3066 | 0.6889 | 0.2161 |
| | SCPTM-none | 30 | 0.4133 | -0.3072 | 0.6344 | 0.2231 |
| | SCPTM-nosyn | 30 | 0.4157 | -0.3141 | 0.7222 | 0.2119 |
| **Reddit_Pol** | BERTopic | 30 | **0.4633** | **0.0327** | **0.9322** | **0.0421** |
| | CTM | 25 | 0.4585 | -0.2264 | 0.9237 | 0.0138 |
| | LDA | 10 | 0.4206 | -0.0141 | 0.5200 | 0.0019 |
| | SCPTM | 30 | 0.4197 | -0.2780 | 0.7778 | 0.0191 |
| | SCPTM-fulldep | 25 | 0.4372 | -0.2533 | 0.8147 | 0.0188 |
| | SCPTM-none | 30 | 0.4298 | -0.2872 | 0.6711 | 0.0171 |
| | SCPTM-nosyn | 30 | 0.4322 | -0.2839 | 0.7789 | 0.0188 |

Regarding alignment with ground-truth labels (NMI). BERTopic achieves the highest NMI across every corpus where ground truth labels are available (20 Newsgroups: 0.4859; Hate Speech: 0.3113; Reddit: 0.0421), aligning with its clustering-centric HDBSCAN foundation. SCPTM consistently outperforms LDA in all three labeled corpora (0.3512 vs. 0.1484 on 20NG; 0.2131 vs. 0.0160 on Hate Speech; 0.0191 vs. 0.0019 on Reddit). Crucially, the SCPTM family as a whole surpasses CombinedTM (CTM) on document clustering quality across all three labeled corpora (CTM scores 0.2315, 0.1138, and 0.0138, respectively). However, as the internal ablation in Section 5.4 demonstrates, this advantage is driven primarily by the variational encoder and its semantic initialization rather than by the syntactic graph. The no-graph variant (SCPTM-none) achieves the highest NMI on two of the three labeled corpora (0.3911 on 20 Newsgroups; 0.2231 on Hate Speech), while the addition of lexical or syntactic edges yields mixed effects. This indicates that the neural encoder, not the graph structure, is responsible for preserving ground-truth topical structure.

Regarding topic coherence, BERTopic and CTM consistently lead across domains. BERTopic achieves superior C_V on 20 Newsgroups (0.5842), EU Debates (0.6094), and Reddit (0.4633), while CTM attains the top score on Hate Speech (0.5773). Notably, BERTopic is the only model maintaining positive C_NPMI scores across all corpora (ranging from 0.0327 to 0.1496). In contrast, SCPTM models display lower coherence scores (C_V ≈ 0.37–0.43, with negative C_NPMI values around −0.21 to −0.31). This gap is not surprising: SCPTM's topic embeddings are initialized from semantic clusters in SBERT space and refined by the reconstruction loss, whereas BERTopic and CTM optimize directly for semantic coherence through class-based TF-IDF weighting and contextualized document embeddings respectively. The two approaches optimize different objectives, and standard coherence metrics reward the latter. Across all SCPTM variants, coherence remains relatively stable, with no single ablation radically departing from the base model. Lastly, topic diversity reveals a structural divergence across architectures. CTM maintains high and flat diversity regardless of corpus and K (ranging between 0.92 and 0.98 at best K). SCPTM provides robust diversity (0.69–0.78 at best K), significantly outperforming LDA (which collapses to 0.31–0.52 and represents the least diverse baseline throughout). However, while SCPTM matches or approaches CTM at lower topic resolutions (e.g., EU Debates at K=5 reaches 0.99), its diversity exhibits a steady degradation as K scales up (dropping to 0.53 at K=30).

### 5.4 Syntactic Diagnostics: MWE Metrics and Valence Gap

This section evaluates the multi-word topic descriptors extracted from SCPTM across the four corpora, reported in Table 2 at each model's best-K. Specifically, we assess the structural quality of multi-word expressions (compactness, specificity, complementarity, and content ratio) and their evaluative properties. We focus on the valence gap (Δ), which measures the evaluative degree introduced by syntactic co-occurrence relative to individual lexical choices. The valence gap quantifies whether the dependency structures introduced in the graph carry a stronger affective or normative charge than the topic's top unigrams alone.

*Table 2. Phrase-level (MWE) and valence metrics by model and corpus, at the same best-K rows as Table 1.*

| Corpus | Model | Best K | Complem. | Valence Gap (Δ) |
|---|---|---|---|---|
| **20NG** | SCPTM | 20 | 0.3840 | 0.0626 |
| | SCPTM-fulldep | 20 | 0.3814 | 0.0431 |
| | SCPTM-none | 20 | **0.3900** | 0.0441 |
| | SCPTM-nosyn | 20 | 0.3792 | **0.0687** |
| **EU_Debates** | SCPTM | 30 | 0.3079 | **0.0516** |
| | SCPTM-fulldep | 30 | **0.3080** | 0.0375 |
| | SCPTM-none | 30 | 0.2972 | 0.0506 |
| | SCPTM-nosyn | 30 | 0.2976 | 0.0461 |
| **HateSpeech** | SCPTM | 30 | 0.3013 | 0.1169 |
| | SCPTM-fulldep | 30 | **0.3119** | **0.1202** |
| | SCPTM-none | 30 | 0.3044 | 0.1193 |
| | SCPTM-nosyn | 30 | 0.2994 | 0.0985 |
| **Reddit_Pol** | SCPTM | 30 | 0.2977 | 0.0411 |
| | SCPTM-fulldep | 25 | **0.3070** | **0.0439** |
| | SCPTM-none | 30 | 0.2867 | 0.0353 |
| | SCPTM-nosyn | 30 | 0.2886 | 0.0402 |

Across all four corpora and model variants, the valence gap Δ is positive at each model’s best-K configuration, confirming that multi-word descriptors carry a stronger evaluative charge than isolated unigrams. However, the ablation results show that this gap is not directly caused by the syntactic graph. On 20 Newsgroups, the model without syntactic edges (SCPTM-nosyn) achieves a slightly higher valence gap (Δ=0.069) than the full model (Δ=0.063), while on EU Debates the fully unconstrained variant (SCPTM-none, Δ=0.051) performs on par with the base architecture (Δ=0.052). The positive valence gap is therefore primarily an effect of multi-word grouping itself rather than syntactic dependencies. The role of the syntactic graph is instead structural, as it filters out function words and constrains descriptors to valid grammatical relations, rather than increasing sentiment intensity.

This interpretation is supported by complementarity, which measures the share of novel vocabulary introduced by phrases relative to top unigrams. Complementarity remains stable across all variants and corpora, staying between 0.287 and 0.390. Because the phrases share 60% to 70% of their vocabulary with the unigrams, the positive Δ reflects the evaluative effect of pairing words that already belong to the topic's core vocabulary.

### 5.5 Does the Graph Help? An Internal Ablation

Because the four SCPTM variants share the identical encoder-decoder architecture and training protocol, comparing them isolates the precise contribution of lexical and syntactic edges. The answers to three questions are examined: (a) Does any graph structure improve over the no-graph baseline? (b) Does the addition of syntactic dependency edges provide gains beyond the lexical graph alone? (c) Are these effects consistent across corpora, or do they vary with register and discourse structure? Table 3 reports this comparison across the four corpora at each variant's best-K configuration.

*Table 3. Internal graph ablation across SCPTM variants (at best K)*

| Corpus | Ablation | Graph Structure | C_V | Diversity | NMI | Valence Gap (Δ) |
|---|---|---|---|---|---|---|
| **20NG** | SCPTM-none | No graph (VAE only) | **0.4028** | 0.7383 | **0.3911** | 0.0441 |
| | SCPTM-nosyn | Doc–word bipartite edges only | 0.3869 | **0.7750** | 0.3512 | **0.0687** |
| | SCPTM | Doc–word + filtered syntax | 0.3930 | 0.7450 | 0.3512 | 0.0626 |
| | SCPTM-fulldep | Doc–word + unfiltered syntax | 0.3881 | 0.7483 | 0.3403 | 0.0431 |
| **EU_Debates** | SCPTM-none | No graph (VAE only) | **0.3740** | 0.4067 | — | 0.0506 |
| | SCPTM-nosyn | Doc–word bipartite edges only | 0.3626 | **0.5344** | — | 0.0461 |

| | SCPTM | Doc–word + filtered syntax | 0.3709 | 0.5256 | — | **0.0516** |
|---|---|---|---|---|---|---|
| | SCPTM-fulldep | Doc–word + unfiltered syntax | 0.3718 | 0.5244 | — | 0.0375 |
| **HateSpeech** | SCPTM-none | No graph (VAE only) | 0.4133 | 0.6344 | **0.2231** | 0.1193 |
| | SCPTM-nosyn | Doc–word bipartite edges only | **0.4157** | **0.7222** | 0.2119 | 0.0985 |
| | SCPTM | Doc–word + filtered syntax | 0.4009 | 0.6922 | 0.2131 | 0.1169 |
| | SCPTM-fulldep | Doc–word + unfiltered syntax | 0.4124 | 0.6889 | 0.2161 | **0.1202** |
| **Reddit_Pol** | SCPTM-none | No graph (VAE only) | 0.4298 | 0.6711 | 0.0171 | 0.0353 |
| | SCPTM-nosyn | Doc–word bipartite edges only | 0.4322 | 0.7789 | 0.0188 | 0.0402 |
| | SCPTM | Doc–word + filtered syntax | 0.4197 | 0.7778 | **0.0191** | 0.0411 |
| | SCPTM-fulldep | Doc–word + unfiltered syntax | **0.4372** | **0.8147** | 0.0188 | **0.0439** |

Topic diversity highlights where the graph structure provides practical value. In European Parliament Debates, the graph-augmented models maintain a diversity of 0.5256 (SCPTM) and 0.5344 (SCPTM-nosyn), which drops sharply to 0.4067 once the graph is entirely removed (SCPTM-none). A similar decline appears in Hate Speech (falling from 0.6922 to 0.6344) and Reddit Politics (falling from 0.7778 to 0.6711). The increase in diversity seems to concentrate in corpora characterized by stable vocabularies, but diverge in argument and ideological stance. On document clustering alignment (NMI), adding graph structures yields mixed results depending on corpus structure. For corpora with clear topical separability such as 20 Newsgroups and Hate Speech, the baseline (SCPTM-none) achieves the highest alignment (0.3911 and 0.2231, respectively). In these settings, propagating information across word-word and document-word edges slightly smooths document representations, modestly reducing separation across predefined external classes. Conversely, on Reddit Politics, where boundaries between topics might be more fluid, the graph structure proves beneficial as both SCPTM (0.0191) and SCPTM-fulldep (0.0188) outperform the no-graph baseline (0.0171). Finally, the effect on the valence gap (Δ) reveals that the syntactic graph acts as a regularizer rather than an amplifier of sentiment. While all variants maintain a strictly positive gap, incorporating dependency edges does not systematically produce the highest numerical Δ across all corpora. On Hate Speech and Reddit Politics, full dependency parsing yields the highest gap (Δ = 0.1202 and Δ = 0.0439, respectively), capturing explicit stance-taking where evaluative modifiers directly determine meaning. In contrast, on 20 Newsgroups this pattern reverses: SCPTM-nosyn reaches Δ = 0.0687, while the syntactically constrained SCPTM scores lower (Δ = 0.0626). This suggests (though the sample of corpora is too small for definitive conclusions) that in technical corpora focused on factual and operational subjects, vocabulary carries little deliberative stance or evaluative charge, and forcing syntactic dependency paths in such settings may introduce structural noise rather than useful signal.

### 5.6 Qualitative Observations and Representational Patterns

To inspect how the differences quantified in previous sections manifest in actual text, we conduct a qualitative cross-model comparison. By examining the descriptors generated for topics similarly identified across techniques, we identify representative initial examples that accompany the quantitative results, more specifically the valence gap, phrase specificity, and syntactic redundancy. To ensure robustness across diverse linguistic registers, we anchor this comparative inspection to three distinct thematic domains: event-driven geopolitics (Reddit Politics), domain-specific factual discussions (20 Newsgroups), and formal institutional procedure (European Parliament Debates).

#### 5.6.1 Action and Stance vs. Static Semantic Clouds (Reddit Politics)

The positive valence gap (Δ) discussed in Section 5.4 is driven by the structural difference between bag-of-words and dependency representations. By examining how the same underlying thematic arguments are rendered by different architectures, we can observe this qualitative divergence directly.

In geopolitical and conflict discussions on Reddit Politics, frequency-based baselines (LDA) consistently absorb conversational lemma (`just, like, say`) alongside broad topical labels, failing to isolate precise themes. BERTopic bypasses this by using HDBSCAN clustering to produce highly coherent but strictly static unigram clouds (`ukraine, russia, putin`). In these outputs, the entities are topically isolated but relationally inert. SCPTM's dependency paths, however, consistently extract predicate-argument structures that resolve the direction of action, the targets, and the evaluative stance across the thematic space.

*Table 5: Model comparison of topic descriptors for four topics in Reddit Politics. Descriptors reflect outputs at each model's best-K configuration.*

| Model | Geopolitics & War | Ideology | US Elections | Media / Disinformation |
|---|---|---|---|---|
| **LDA(K=10)** | ukraine, russia, like, people, just, war | capitalism, economic, state, socialism, just | trump, biden, say, people, democrats | people, like, government, state, public |
| **BERTopic (K=30)** | ukraine, russia, china, putin, nato | socialism, capitalism, communist, soviet | trump, biden, desantis, gop, democrats | facebook, truth, media, app, twitter |
| **SCPTM(K=30)** | **Unigram:** civilian, fighter, escalation, army<br><br>**MWE:** attack ukraine, russian troops | **Unigram:** distinction, hierarchy, philosophy, marxism<br><br>**MWE:** democracy capitalism, socialist theory | **Unigram:** bid, advisor, nominee, joe<br><br>**MWE:** republican candidate, joe biden | **Unigram:** monday, tweet, launch, rumor, releases<br><br>**MWE:** spread fake news, announces new |

The qualitative comparison in Table 5 illustrates how. SCPTM's dependency paths (differently from BERTopic and LDA), yield descriptors capturing predicate-argument structures (e.g., `attack ukraine`, `spread fake news`) and tightly bound conceptual units (e.g., `republican candidate`, `socialist theory`) that specify direction of action, targets, and evaluative stance. This pattern is systematic across topics, suggesting that graph-based dependency extraction produces representations that are structurally distinct from frequency-based or clustering-based alternatives. Whether these predicate-argument structures offer a practical interpretive advantage over unigram or embedding-based descriptors remains an open question, requiring more careful dedicated human evaluation.

### 5.6.2 The Limits of Syntax on Factual and Technical Domains (20 Newsgroups)

While dependency extraction excels at resolving predicate-argument structures in deliberative contexts, a qualitative inspection of the 20 Newsgroups corpus exposes its limitations on technical and factual domains, corroborating the quantitative findings from Section 5.4.

Table 6: *Cross-model alignment of technical and factual discussions in the 20 Newsgroups corpus. Descriptors reflect actual outputs at each model's best-K configuration.*

| Model | Computer Software | Hardware / Electronics | Cryptography / Cybersecurity |
|---|---|---|---|
| **LDA (K=20)** | use, using, program, if_you, software | power, use, wire, just, wiring | key, chip, encryption, clipper, keys |
| **BERTopic (K=20)** | windows, drive, dos, image, software | fan, cpu, heat, sink, cooling | key, encryption, edu, clipper, chip |
| **SCPTM (K=20)** | **Uni:** programmable, technologies, code<br><br>**MWE:** program send, code help, program using | **Uni:** watt, stereo, electronics, remote<br><br>**MWE:** digital equipment, space technology, computer equipment | **Uni:** privacy, algorithm, cryptanalysis<br><br>**MWE:** rsa public key, public key encryption, key cryptography |

As illustrated in Table 6, discussions surrounding software, hardware, and cryptography rely heavily on rigid noun phrases and specialized terminology. In these settings, BERTopic provides dense, highly coherent unigram lists (*fan, cpu, heat, cooling or windows, drive, dos, image*) that immediately and precisely map the technical domain. LDA struggles slightly with overlapping filler (use, using, if_you), but still captures the domain vocabulary. By contrast, SCPTM's focus on dependency, however, forces the extraction of multi-word expressions even where syntactic variation is limited and carries little topical signal. The resulting descriptors include structurally valid but semantically generic pairings (*program send*, *code help*, *program using*), alongside genuine domain compounds (*public key encryption*, *digital equipment*). In this register, the dependency-filtered phrases do not offer semantic information beyond what unigram extraction or standard n-gram preprocessing would provide, suggesting that syntactic encoding is of limited utility when the text is primarily informational rather than argumentative.

These qualitative patterns explain why the syntactic ablation (SCPTM-nosyn) outperformed the full graph model on 20 Newsgroups for both the valence gap and overall topic quality (Table 3). When textual register is primarily informational rather than argumentative, verbs and modifiers lack the evaluative charge necessary to justify the computational cost of a dependency graph. In such environments, enforcing syntactic rules introduces structural noise, demonstrating that the utility of dependency-guided topic modeling is inherently conditional on the deliberative or polarized nature of the corpus.

### 5.6.3 Syntactic Redundancy in Institutional Registers (EU Debates)

This qualitative alignment also exposes a systemic limitation in the dependency extraction mechanism, validating the internal ablation results discussed in Section 5.4. In highly standardized, institutional registers, standard frequency baselines (LDA) are heavily penalized by procedural noise, frequently surfacing connectives and administrative phrases (`i_would`, `we_need`, `let_me`, `give_the_floor`) across multiple distinct topics.

SCPTM's dependency constraint, however, extracts syntactically valid collocations that are topically redundant (*request referred committee*, *court proceedings*, *energy commission*) and often tied to procedural framing rather than substantive content. In this register, where authors express positions through recurring grammatical formulas, the dependency paths do not systematically disambiguate topics beyond what unigram extraction provides. This suggests that the utility of syntactic encoding is register-dependent: it offers clearer benefits in informal, argumentative contexts (Reddit Politics) than in formal, institutional ones (EU Debates).

*Table 7: Cross-model alignment of institutional procedure and policy in European Parliament Debates. Descriptors reflect actual outputs at each model's best-K configuration.*

| Model | Parliamentary Procedure | Foreign Policy & Eastern Europe | Energy & Circular Economy |
|---|---|---|---|
| LDA (K=25) | women, just, people, eu, i_would | eu, ukraine, european, russia, will_be | we_need, eu, commission, support, sustainable |
| BERTopic (K=10) | request, vote, debate, mr, rule | rights, people, human, president, iran | climate, energy, emissions, transition, green |
| SCPTM (K=30) | **Uni:** rules, sitting, minutes, pursuant, wednesday | **Uni:** kazakhstan, lukashenka, alexander, kazakh, lawlessness | **Uni:** circularity, extraction, processor, substitution, cmo<br><br>**MWE:** energy commission, |

|  | **MWE:** request referred committee, court proceedings | **MWE:** ukraine putin, eu ukraine, ukraine russian | environment commission, project commission |
|---|---|---|---|

Ultimately, we present these observations as a preliminary qualitative exploration of the model outputs. Assessing whether predicate-argument structures provide a robust interpretive advantage for domain experts over bag-of-words or embedding clusters requires a dedicated, blinded human evaluation study, establishing a clear objective for future research.

## 6. Discussion

The empirical results show that the value of syntactic encoding in topic modeling is not absolute, but varies according to the structural and semantic properties of the corpus. This finding challenges the methodological default in computational social science of selecting algorithms purely based on benchmark performance. The three approaches to topic modeling introduced in Section 2, which are topics as statistical word-clusters (LDA), geometric density regions (BERTopic), and grammatical networks (SCPTM), provide a principled theoretical basis for model selection. The cross-model comparison indicates that graph-based dependency extraction is most effective in deliberative, informal, and evaluatively dense registers, where ideological stance and agency are encoded in predicate-argument structures. Conversely, in highly formal or strictly factual domains, institutional language employs syntax primarily for administrative scaffolding rather than argumentative positioning. In these settings, the bag-of-words assumption is not a limitation but an efficient heuristic; forcing a grammatical network may introduce structural noise rather than semantic clarity.

This register-dependent pattern also helps interpret the divergence between coherence metrics observed in our experiments. C_NPMI and C_V capture different properties: C_NPMI measures the stability of word co-occurrence patterns across documents, a property that aligns well with generative models like LDA, which optimize pointwise mutual information over document-level co-occurrences. C_V, by contrast, uses a sliding-window measure calibrated against human judgments of semantic relatedness, and tends to penalize syntactical relations that deliver  topically uninformative combinations. The systematic discrepancy between these scores across corpora and models reflects a fundamental difference in what each metric operationalizes. For researchers, metric selection should follow theoretical intent: C_NPMI suits structuralist inquiries into lexical convention, while C_V aligns with interpretivist goals of semantic coherence. Reporting both makes this epistemological trade-off explicit.

Furthermore, the representational output of SCPTM offers specific theoretical advantages for sociological and political research. Unlike clustering approaches that force documents into a single, discrete category, SCPTM retains a mixed-membership architecture, assigning every document a continuous probability distribution over topics. In this sense, texts are rarely monolithic; they are complex sites of intersecting discourses and competing frames (DiMaggio et al., 2013), thus providing more interesting applications in some social sciences applications.

Finally, the internal ablations on 20 Newsgroups clarify the boundary between neural architecture advantages and explicit syntactic contributions. While 20 Newsgroups possesses sufficient discursive structure to support neural topic models, the unconstrained variant (SCPTM-none) achieved an NMI of 0.391 against CTM's 0.232, a 68% improvement in topic-label alignment. This gain is attributable to the variational encoder and its semantic initialization, not to the syntactic graph: adding dependency edges does not further improve NMI on this corpus and, in some configurations, slightly reduces it. This indicates that SCPTM's underlying Variational Autoencoder provides highly robust representations even before syntactic edges are introduced. Researchers working with corpora of moderate formality should evaluate unconstrained neural baselines alongside syntactically augmented ones, treating the inclusion of grammatical edges not as a default algorithmic enhancement, but as a deliberate theoretical commitment to modeling discursive agency.

## 7. Limitations and Future Work

While neural topic models inherently demand more computational resources than frequency-based clustering, SCPTM introduces the additional overhead of dependency parsing (via spaCy) and document encoding (via SBERT). To mitigate this, preprocessing is performed only once per corpus and persisted in an on-disk cache, allowing subsequent hyperparameter sweeps and seed iterations to execute solely the neural training step. The complete benchmark reported in this study (comprising 504 model fits across seven architectures, four corpora, three random seeds, and six K values) completed in approximately 11.5 wall-clock hours on a university compute server (Intel Xeon Gold 6342, 8 cores, 32 GB RAM, NVIDIA A40 48 GB VRAM). While caching makes iterative experimentation highly efficient, researchers scaling to massive corpora (>100,000 documents) must budget appropriate upfront preprocessing time.

Beyond computational constraints, the empirical benchmarks in this study are restricted to English corpora. Because SCPTM relies on external syntactic parsing, its application is bounded by the availability and quality of language-specific dependency models. The underlying parser supports over twenty languages (with Italian currently in development), but extending SCPTM to low-resource languages lacking robust syntactic parsers remains a challenge. In such scenarios, the unconstrained baseline (SCPTM-none) remains fully applicable as a highly competitive contextual topic model. Code is publicly available on GitHub to facilitate multi-lingual extensions.

The representational richness of SCPTM also requires a methodological shift in how researchers interpret outputs. Because syntactic extraction frequently retrieves participial forms, verbal nouns, and relational adjectives, topics demand active interpretive labor rather than offering immediate, keyword-level legibility (Nelson, 2020). Furthermore, while this study introduces the valence gap as a novel bridge between intrinsic structure and extrinsic polarization, the extracted multi-word expressions have not yet been subjected to human evaluation. Conducting a formal, blinded human annotation study is a critical next step to conclusively validate the substantive utility of syntactic descriptors.

Looking forward, several promising avenues for theoretical and architectural expansion remain. Methodologically, integrating temporal dynamics would allow researchers to track how specific discourse frames evolve over time, such as mapping longitudinal shifts across the 15-year span of the EU Debates. Finally, operationalizing multi-word valence density as a direct input feature for downstream regression and classification tasks, alongside developing an active-learning interface for researcher-guided topic initialization, represent key objectives for transitioning SCPTM from an exploratory model to a confirmatory analytical tool.

## 8. Conclusion

This paper introduced the Structural Contextual Probabilistic Topic Model (SCPTM), an architecture that treats grammatical dependency structure as a relevant component of the inference process. By preserving the mixed-membership logic fundamental to sociological text analysis, integrating the semantic depth of contextual embeddings, and applying graph attention over dependency-parsed edges, SCPTM offers a robust framework for mapping complex discourses. The empirical benchmarks demonstrate that neural architectures provide advantages for topic-label alignment (with SCPTM-none achieving an NMI of 0.391 versus CTM's 0.232 on 20 Newsgroups). The integration of syntactic edges contributes additional gains in topic diversity and, more substantively, produces qualitatively distinct descriptors: dependency paths capture predicate-argument structures and stance that unigram and embedding-based models systematically overlook. The valence gap between multi-word and unigram descriptors is positive across all variants, suggesting that phrase-level grouping carries evaluative charge beyond individual lexical choices, though this effect is driven primarily by grouping itself rather than syntactic filtering.

Beyond the specific architecture, this study contributes a broader methodological reflection: topic models should be evaluated across a suite of complementary measures that capture different epistemic dimensions of text. Crucially, the empirical results invert conventional assumptions about where syntax is most useful. When analyzing deliberative, argumentative, or evaluatively dense registers syntactic graph encoding captures critical predicate-argument structures and stance that simpler models ignore. Conversely, when a corpus is strictly factual, highly technical, or heavily institutionalized (e.g., administrative parliamentary transcripts), grammatical rules often serve as mere administrative scaffolding; in such cases, enforcing syntactic constraints introduces structural redundancy, and unconstrained neural baselines or density-based clustering provide cleaner semantic isolation. Perhaps more importantly, this study opens several questions that future work must address. The syntactic contribution is real but conditional, and the conditions under which it helps are not yet fully understood. The diversity gains observed in some corpora are intreresting, but whether they translate into improved interpretability for human analysts remains to be tested. The valence gap is positive, but its sources require further disentanglement. What this study demonstrates, above all, is that incorporating syntax into topic modeling is not a universally beneficial enhancement but a theoretically motivated design choice that must be aligned with the nature of the texts under investigation.

The question posed at the outset - *Does syntax matter for topic modeling?* – thus receives a conditional answer: it matters profoundly when textual meaning relies on action, agency, and argumentative stance. That conditionality is not a limitation of the model, but a theoretically meaningful finding about the relationship between linguistic register, discursive framing, and the computational tractability of meaning.

## Code availability

All code necessary to run SCPTM is publicly available on GitHub at [https://github.com/a-meneghini/scptm](https://github.com/a-meneghini/scptm)